\documentclass[runningheads]{llncs}
\usepackage[T1]{fontenc}
\usepackage{graphicx,verbatim}
\usepackage{xspace}
\usepackage{gensymb}
\usepackage{algorithm}
\usepackage{multirow}
\usepackage{algpseudocode}
\usepackage{hyperref}
\usepackage{amsmath}
\usepackage{xcolor}
\usepackage{amssymb}
\usepackage{color}

\newcommand{\MetaSeg}{MetaSeg\xspace}

\newcommand{\dataset}{OASIS-MRI\xspace}
\newcommand{\vect}[1]{\mathbf{#1}}
\newcommand{\vx}{\vect{x}} 
 
\newcommand{\va}{\vect{a}} 
\newcommand{\vb}{\vect{b}} 

\begin{document}

\title{Fit Pixels, Get Labels: Meta-Learned Implicit Networks for Image Segmentation}

\author{Kushal Vyas \and
Ashok Veeraraghavan \and
Guha Balakrishnan}
\authorrunning{K.Vyas et al.}
%
\institute{Rice University, Houston\\
\email{\{kvyas,vashok,guha\}@rice.edu}}

\maketitle              

\begin{abstract}
Implicit neural representations (INRs) have achieved remarkable successes in learning expressive yet compact signal representations. However, they are not naturally amenable to predictive tasks such as segmentation, where they must learn semantic structures over a distribution of signals. In this study, we introduce \MetaSeg, a meta-learning framework to train INRs for medical image segmentation. \MetaSeg uses an underlying INR that simultaneously predicts per pixel intensity values and class labels. It then uses a meta-learning procedure to find optimal initial parameters for this INR over a training dataset of images and segmentation maps, such that the INR can simply be fine-tuned to fit pixels of an unseen test image, and automatically decode its class labels. We evaluated \MetaSeg on 2D and 3D brain MRI segmentation tasks and report Dice scores comparable to commonly used U-Net models, but with $90\%$ fewer parameters. \MetaSeg offers a fresh, scalable alternative to traditional resource-heavy architectures such as U-Nets and vision transformers for medical image segmentation. Our project is available \href{https://kushalvyas.github.io/metaseg.html}{here}.

\end{abstract}

\section{Introduction}
Implicit neural representations (INRs) have demonstrated impressive performance in encoding complex natural signals. A typical INR $f_{\theta}(\cdot)$ is a multilayer perceptron (MLP) with parameters $\theta$ that maps a coordinate 
$\mathbf{x} \in \mathrm{R}^d$ to signal value $I(\mathbf{x}) \in \mathrm{R}^D$ and is 
iteratively fit 
to a specific signal. 
Because they offer continuous signal representations, excellent reconstruction performance, and implicit signal priors, INRs have gained significant attention in the computer vision community for compactly modeling large signals\cite{acorn,miner,wire}, and in the medical imaging community for inverse imaging tasks such as accelerated MRI~\cite{inrfusion}, and sparse-view CT reconstruction~\cite{mlr_inr_ct,naf}. Despite these successes, learned INR representations are highly specific to a given signal and to the way its parameters are initialized. As a result, unlike architectures such as U-Nets~\cite{unet} and vision transformers~\cite{vit_medical}, INR-produced features lack structural or semantic coherence, making them less used for tasks requiring learning over an image distribution. 

Results from recent studies suggest that this shortcoming of INRs may be fixable. In particular, by fitting an INR across multiple images starting from the same parameter intialization, the final parameters exhibit clear semantic, structural properties for datasets such as faces or MRI scans~\cite{functa,learnit,strainer}. And as a result, these learned initializations allow the INR to fit unseen (test) images rapidly with far fewer gradient updates while learning more generalizable features~\cite{spatial_functa,learnit,strainer}. In this study, we leverage this insight to develop an INR fitting strategy for medical image segmentation. In particular, we present the surprising discovery that an INR carefully optimized to fit many training pairs of images and associated segmentation maps can predict a segmentation map for an unseen test image \emph{when simply fine-tuned to reconstruct the image's pixels.}

\begin{figure}[t!]
    \centering
    \includegraphics[width=\linewidth]{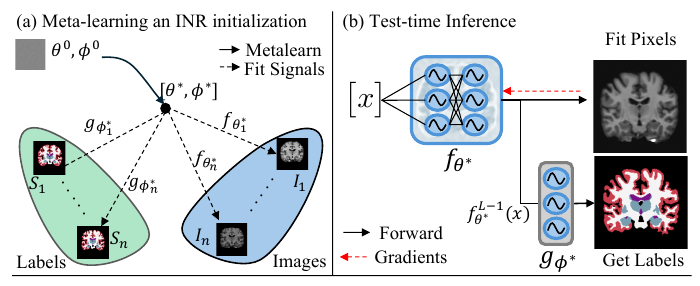}
    \caption{\textbf{Overview of \MetaSeg.} (a) We use a meta-learning framework to learn optimal initial parameters $\theta^*, \phi^*$ for an INR consisting of an $L$-layer reconstruction network $f_\theta(\cdot)$ and shallow segmentation head $g_\phi(\cdot)$. (b) At test time, optimally initialized INR $f_{\theta^*}$ is iteratively fit to the pixels of an unseen test scan. After convergence, the penultimate features $f_{\theta^*}^{L-1}(x)$ are fed as input to the segmentation head $g_\phi^*(\cdot)$ to predict per-pixel class labels.}
    \label{fig:pipeline}
\end{figure}

We propose \MetaSeg, a novel meta-learning~\cite{maml} INR framework that performs this strategy. \MetaSeg uses an INR that simultaneously predicts both image pixels and segmentation labels given a coordinate (Fig.~\ref{fig:pipeline}b), and aims to find an optimal setting of the INR parameters ($\theta$ and $\phi$), such that the network can be quickly fine-tuned to fit any image-mask pair starting from these parameters. \MetaSeg uses a nested, meta-learning procedure (Fig.~\ref{fig:pipeline}a). In the inner optimization, the INR is fit to reconstruct a randomly selected batch of training image and segmentation mask pairs. The outer (meta) optimization applies a gradient descent step to the current values of $\theta$ and $\phi$ based on the converged inner parameters. These steps iterate to converge to an initialization that works well for both image reconstruction and segmentation objectives from the given image distribution. At test time, we simply fit the INR initialized with the learned optimal parameters to the test image (without needing any segmentation information), generating viable segmentation masks in just $2$ gradient descent updates and near state-of-the-art quality masks after $100$ updates.

\MetaSeg draws inspiration from Neural Implicit Segmentation Functions (NISFs)~\cite{nisf}, which also use an INR to predict pixel and segmentation values. However, instead of forcing the INR to learn image-label correlations, NISF takes an additional latent vector as input to represent an image's semantic content, which must then be optimized at test time to produce a mask. Due to this strategy, NISF must apply several regularization losses on INR parameters and the latent space to succeed. In contrast, \MetaSeg offers a much simpler training methodology based on meta-learning, only requiring reconstruction and segmentation losses during training and only a reconstruction loss during testing.

We evaluated \MetaSeg on 2D and 3D magnetic resonance imaging (MRI) segmentation tasks using the \dataset~\cite{oasis1,oasis2} dataset. We first find that \MetaSeg achieves similar performance for 2D (Dice score: $0.93$) and 3D (Dice score: $0.91$) MRI segmentation tasks to widely used U-Net baselines~\cite{unet_brain,segresnet} with $90\%$ fewer parameters, and significantly outperforms NISF~\cite{nisf}. Second, we show that \MetaSeg performs well for fine-grained segmentation tasks,
accurately segmenting image regions which may be sparse and span only a few pixels.
Third, we show that when trained on low-resolution scans, \MetaSeg fares better on high-resolution scans than U-Nets, likely due to its continuous representation. Finally, we qualitatively demonstrate through principal component analysis (PCA) visualizations that \MetaSeg's learned initial parameters encode semantic information correlated with ground truth segmentations.

\section{Methods}
\label{sec:methods}
We assume a given dataset $\mathcal{D} = \{I_j, S_j\}_{j=1}^N$ of $N$ subjects, where $I_j$ and $S_j$ are a 
$d$-dimensional image scan and segmentation map, respectively 
($d=2$ and $3$ in our experiments). At  location $x \in \mathbb{R}^d$, we have $I_j(x) \in \mathbb{R}^D$ and 
$S_j(x) \in \{0,1\}^{|\mathcal{C}|}$, where 
$D$ is the number of image channels, and $C$ is the number of classes. 

We propose the \MetaSeg framework to estimate $\hat{S}$ for an unseen scan $I$ at test time. \MetaSeg consists of an INR (Fig.~\ref{fig:pipeline}b) with an $L$-layer MLP 
$f_{\theta}: \mathrm{R}^d \rightarrow \mathrm{R}^D$ that predicts $\hat{I}(\vx)$
given coordinate $\vx$, and a segmentation head $g_{\phi}: \mathrm{R}^{h} \rightarrow \mathrm{R}^{C}$ that predicts $C$ class probabilities given the penultimate features, of dimensionality $h$, computed by $f_\theta$: $f^{L-1}_{\theta}(\vx)$. At test time, we aim to fit $f_\theta(\cdot)$ on $I$ for $T_f$ steps, and then simply compute $\hat{S}(\vx) = g_\phi(f^{L-1}_{\theta}(\vx))$ in one feed-forward operation. 

In Sec.~\ref{sec:init_meta}, we describe \MetaSeg's strategy for learning optimal values $\theta^*, \phi^*$ over $\mathcal{D}$, such that $f_{\theta^*}(\cdot)$ and $g_{\phi^*}(\cdot)$ (networks initialized with $\theta^*$ and $\phi^*$) may be easily fine-tuned to reconstruct and segment a scan from the distribution. In Sec.~\ref{sec:inference}, we describe how the INR initialized with these parameters may be used at inference time to estimate a segmentation map for a scan.

\subsection{Training: Meta-Learning and Segmentation Head Optimization} 
\label{sec:init_meta}
\MetaSeg uses a MAML\cite{maml} meta-learning strategy to learn optimal parameters $\theta^*,\phi^*$, consisting of a nested optimization with \emph{inner} and \emph{outer} (meta) routines (Fig.~\ref{fig:pipeline}a). Starting with initial random values $\theta^0$ and $\phi^0$ at iteration $t=0$, the outer routine takes $T_o$ gradient descent steps to move the parameters to generalizable values. The gradient at each outer step is based on the results of the inner routine, which fits $f_{\theta^t}(\cdot)$ and $g_{\phi^t}(\cdot)$ for $T_i$ steps on one data pair $(I_j, S_j)$. The outer routine uses the final inner parameter values as a gradient signal to update $\theta^{t},\phi^{t}$. To avoid a noisy gradient due to overfitting to example $j$, we set $T_i$ to a small value in practice (e.g., $T_i=2$).

In the inner routine, at time step $t$, we optimize both $f_{\theta^t}(\cdot)$ and $g_{\phi^t}(\cdot)$ for a single subject $j$ with a loss function:
\begin{equation}
    \mathcal{L}_{inner}(I_j, \hat{I}_j, S_j, \hat{S}_j) = \sum_\vx \mathcal{L}_{recon}(I_j(\vx), \hat{I}_j(\vx)) + \mathcal{L}_{cls}(S_j(\vx), \hat{S}_j(\vx)),
\end{equation} 
where $\mathcal{L}_{recon}(\va, \vb) = \mid\mid \va - \vb \mid\mid^2_2$ is a per-pixel reconstruction loss, and 

\begin{equation}
\mathcal{L}_{cls}(\va, \vb) = \sum_{c=1}^C - ( 1 - \vb(c)){^\gamma} \cdot \delta_{c,\va} \log(\vb(c))
\label{eq:focal}
\end{equation}
is a per-pixel multiclass focal classification loss~\cite{focal_loss}, {where $\gamma$ is a hyperparameter. We use a focal loss to account for heavy class imbalance across pixels.} In the outer step, we make a gradient descent update:
\begin{equation}
    \label{eq:l_outer_loss}
    [\theta^{t+1}, \phi^{t+1}] \leftarrow [\theta^{t}, \phi^{t}] - \beta \nabla_{[\theta^{t}, \phi^{t}]}  \big{(}\mathcal{L}_{inner}(I_j, \hat{I}_j, S_j, \hat{S}_j) \big{)}, 
\end{equation}
where gradient $\nabla_{[\theta^t, \phi^t]} (\cdot)$ computes the difference between the converged parameters ($\theta^t_j$, $\phi^t_j$) of the inner optimization and current parameters $\theta^t, \phi^t$. After $T_o$ outer gradient descent steps, we obtain parameters $\theta^{T_o}$ and $\phi^{T_o}$. 

We freeze $\theta^* = \theta^{T_o}$, but further optimize $\phi^{T_o}$ such that when, at test time, $f_{\theta^*}(\cdot)$ is fit for $T_f$ iterations on any random scan, we can obtain accurate segmentation predictions, in one step.

To do so, we first fit $f_{\theta^*}(\cdot)$ separately on each training scan $I_j$ for $T_f$ iterations and populate a dataset of learned scan-specific features and segmentation maps, $\{f_{\theta^*_j}^{L-1}(\vx), S_j(\vx)\}_{j=1}^N$. We then globally optimize $g_\phi(\cdot)$ by minimizing the loss function $\mathcal{L}_{seg}$:
\begin{equation}
\mathcal{L}_{seg}(\mathcal{D}) = \sum_{\vx} \sum_j \mathcal{L}_{cls}(S_j(\vx), g_{\phi}(f_{\theta^*_j}^{L-1}(\vx))). 
\end{equation}
We freeze the converged parameters as $\phi^*$.

\subsection{Inference: Fit Pixels, Get Labels} 
\label{sec:inference}
At inference time, we assume a given (unseen) scan $I$. We fit $f_{\theta^*}(\cdot)$ to $I$ for $T_f$ iterations, solely optimizing the pixelwise reconstruction loss $\mathcal{L}_{recon}$. Upon completion, we compute $\hat{S}(\vx) = g_{\phi^*}(f^{L-1}_{\theta^*}(\vx))$, and apply softmax and argmax operations to $\hat{S}(\vx)$ to obtain the predicted class per pixel.

\begin{figure}[t!]
    \centering
    \includegraphics[width=\linewidth]{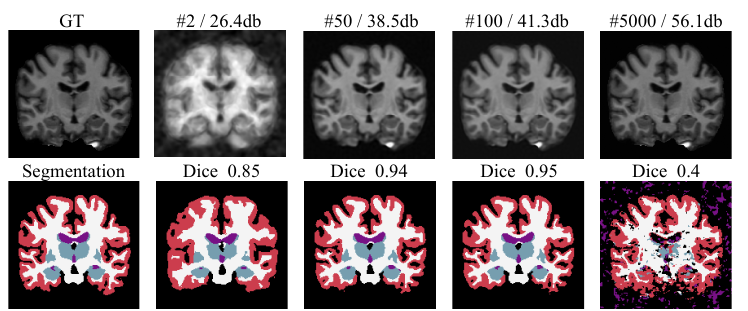}
    \caption{\textbf{Example progression of reconstruction and segmentation performance as a \MetaSeg INR fits the pixels of a test image.} We report PSNR (top) and Dice (bottom) scores. \MetaSeg's initialization leads to rapid convergence, with a Dice score of 0.85 emerge as early as $T_f=2$ iterations. It achieves optimal segmentation (Dice=$0.95$, PSNR=$41.3$) at $T_f=100$ iterations, and eventually declines as it overfits to the image pixels at $T_f = 5000$ iterations.
    }
    \label{fig:meta_progress}
\end{figure}

\section{Experiments and Results}
We evaluated \MetaSeg on the tasks of 2D and 3D MRI image segmentation. We used 414 T1 brain MRI scans from the OASIS-MRI~\cite{oasis1,oasis2} dataset, randomly split into 214 training, 100 validation, and 100 test scans. \dataset provides 5 coarse (background, white matter, gray matter, cortex, and cerebrospinal fluid), 24 fine-grained segmentation labels for 2D slices, and 35 segmentation labels for full 3D MRI volumes. We used aligned coronal sections provided in \dataset for 2D segmentation and aligned 3D volumes for 3D segmentation. We normalized all images and volumes to have intensity in $\begin{bmatrix}0,1\end{bmatrix}$.  We padded and resized images to resolution $192 \times 192$, and cropped 3D volumes to $160\times160\times200$ followed by $2\times$ downsampling to yield $80\times 80 \times 100$ resolution scans.

\textbf{Implementation.} For all experiments, we used a SIREN~\cite{siren} INR with $L=[6,5,5]$ layers, width $h=[128,512,256]$, and $w_0=30$. We set the segmentation head $g_\phi(\cdot)$ as one fully connected layer with Leaky-ReLU activation, followed by a linear layer with $C$ output values. We used the Adam~\cite{adam} optimizer with a learning rate of $10^{-4}$ for inner and outer meta-learning optimizations. When tuning the segmentation head for 5 classes, we reduced the learning rate to $5e-5$. We set $T_i=2$ and $T_f=100$, and ran the outer loop for $T_o=10$ epochs over the training data while validating the performance of optimized parameters $\theta^t,\phi^t$ every $50$ steps. For 2D segmentation with 5 classes, we set the focal loss hyperparameter $\gamma=1.0$, and for 24 output classes we set $\gamma=2.0$. For 3D MRI volumes which exhibit high background class imbalance, we set $\gamma=3.0$ and further scaled the reconstruction loss by $0.1$ for background pixels.  

\textbf{Baselines:} We adopt a widely used U-Net~\cite{unet} baseline for 2D MRIs~\cite{unet_brain} and SegResNet~\cite{segresnet} from the Monai~\cite{monai} package for 3D MRIs. We trained all models using a learning rate of $1e-3$ until the validation loss saturated. We also compare against NISF~\cite{nisf}, the most closely related INR method in the literature.

\begin{table}[t!]
    \centering
    \caption{\textbf{Quantitative performance of \MetaSeg and baseline models on 2D and 3D MRI segmentation tasks.} \MetaSeg achieves comparable Dice scores to popular U-Net baselines with $90\%$ fewer parameters.}
    \label{tab:results_baselines}
    \begin{tabular}{|c|c|c|c|c|}
    \hline
    Task & Num. classes & Model  &  Dice Score $\uparrow$ & Num. parameters $\downarrow$\\
   \hline
       \multirow{4}{*}{2D MRI Segmentation} & \multirow{2}{*}{5} & U-Net~\cite{unet_brain} & $0.96 \pm 0.008$ & 7.7M\\
       & & \MetaSeg & $0.93 \pm 0.012$ & 83K\\
       \cline{2-4}
       & \multirow{2}{*}{24} & U-Net~\cite{unet_brain} & $0.84 \pm 0.097$ & 7.7M \\
       & & \MetaSeg & $0.86 + 0.032$ & 1.06M\\
       \hline
       \multirow{3}{*}{3D MRI Segmentation} & \multirow{3}{*}{5} & SegResNet~\cite{segresnet} & $0.95 \pm 0.006$ & 4.7M\\
       & & NISF~\cite{nisf} & $0.81 \pm 0.007$ & 293K\\
       &  & \MetaSeg & $0.91 \pm 0.011$ & 330K \\
         \hline
    \end{tabular}
\end{table}

\begin{figure}[t!]
    \centering
    \includegraphics[width=\linewidth]{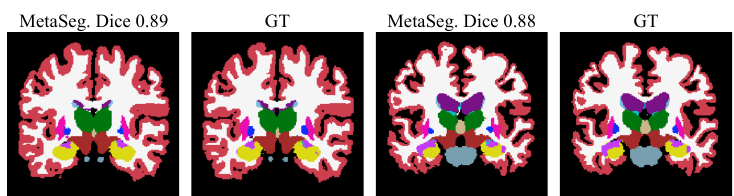}
    \caption{ \textbf{Coarse and fine-grained 2D brain MRI segmentation results with 24 classes, for two subjects.} \MetaSeg accurately segments structures and is robust to high variations across subjects for structures such as ventricles (depicted in purple), brain stem (gray), and hippocampus (yellow). \MetaSeg also adapts well to structures such as the cerebral cortex (red) which are not compact and localized in space.}
    \label{fig:results_2d_seg}
\end{figure}

\subsection{Results}

Table~\ref{tab:results_baselines} presents average Dice\cite{dice} scores with standard deviations for \MetaSeg and baseline methods for both 2D and 3D cases. For 2D, \MetaSeg achieves an excellent Dice score of $0.93 \pm 0.012$ for 5 classes, comparable to that achieved by the U-Net, but with $90\%$ fewer parameters. Interestingly, as Fig.~\ref{fig:meta_progress} illustrates, \MetaSeg fits a signal remarkably fast, yielding a viable segmentation in just $2$ update steps to $f_\theta(\cdot)$ (Dice: $0.85$), and reaching peak performance at $T_f$ steps. If fit further, \MetaSeg's segmentation accuracy eventually falls off as parameters overfit to the particular scan. \MetaSeg also performs well on the finer-grained segmentation task with 24 classes, achieving a Dice score of $0.86 \pm 0.032$ and outperforming the baseline U-Net. As shown in Fig.~\ref{fig:results_2d_seg}, \MetaSeg provides consistently accurate segmentation results for the ventricles and brain stem, which have drastic variations in size and area across examples. 

\begin{figure}[t!]
    \centering
    \includegraphics[width=\linewidth]{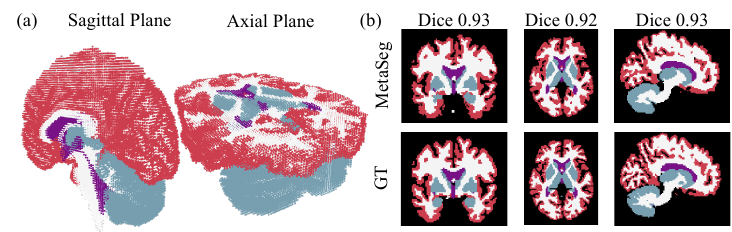}
    \caption{\textbf{\MetaSeg is capable of accurately segmenting 3D MRI volumes.} (a) Various cross sections of one learned volume. (b) Renderings of intermediate coronal, axial, and sagittal planes, with high agreement (Dice$\approx 0.93$) with ground truth.}
    \label{fig:brain_3d}
\end{figure}

\MetaSeg also performs well for 3D segmentation, achieving an average Dice score of $0.91 \pm 0.011$ with $90\%$ fewer parameters than SegResNet, and significantly outperforms NISF. Fig.~\ref{fig:brain_3d} depicts rendered coronal, sagittal, and axial planes from the segmented volume showing high agreement with ground truth segmentation planes, also demonstrating that \MetaSeg-encoded volumes can be readily queried at any viewing plane. We also explored whether \MetaSeg can generate high resolution ($2\times$) 3D segmentation maps when only trained on low resolution scans, due to its underlying continuous representation. \MetaSeg achieved a Dice score of $0.78 \pm 0.011$ on this task, while SegResNet~\cite{segresnet} achieved a Dice score of $0.73 \pm 0.019$. Hence, while there is a significant drop in performance in this scenario, \MetaSeg still outperforms U-Nets.

Finally, we conducted an ablation study on parameter initialization strategy for 2D segmentation, with results reported in Table~\ref{tab:init_ablation}. \MetaSeg performs significantly better than INRs initialized with random, fixed, and image-only meta-learning strategies. The latter result, in particular, demonstrates the benefit of learning structural priors with joint segmentation supervision.

\begin{figure}[t!]
    \centering
    \includegraphics[width=\linewidth]{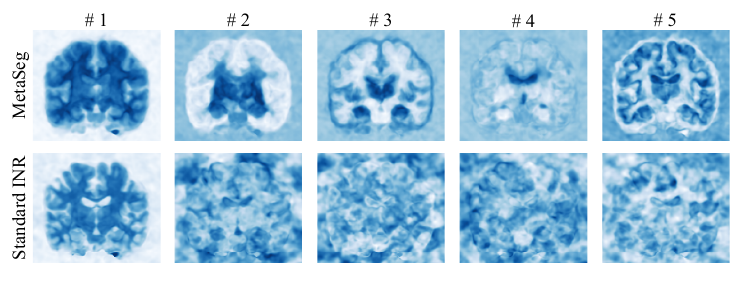}
    \caption{ \textbf{PCA visualization of penultimate features learned by \MetaSeg and a standard INR for a 2D test scan.} We performed PCA separately on the features returned by each INR across spatial coordinates. We see a strong correlation for \MetaSeg features with anatomical structures. For example, component \#2 approximately resembles inner brain regions, \#3 resembles regions like hippocampus and basal ganglia, \#4 resembles ventricles, and \#5 captures the cerebral cortex. On the contrary, a standard INR yields seemingly random features.}
    \label{fig:pca}
\end{figure}

\begin{table}[h!]
\centering
\caption{\textbf{Ablation studies on INR initialization strategies for 2D segmentation.} \MetaSeg's approach of jointly learning initial INR and segmentation parameters outperforms random, fixed, and image-only meta-learning initialization strategies.}
\label{tab:init_ablation}
\begin{tabular}{|c|c|c|c|c|}
\hline
 Init. Strategy &  Random & Fixed & Meta-learn, Image Only & \MetaSeg \\
\hline
Dice scores & $0.30 \pm 0.057$ & $0.53 \pm 0.1 $ & $0.81 \pm 0.033$ & $\mathbf{0.93 \pm 0.012}$\\
\hline
\end{tabular}
\end{table}

\section{Discussion and Conclusion}
Results demonstrate that \MetaSeg achieves strong performance on both 2D and 3D MRI segmentation tasks, with comparable Dice scores to common U-Net baselines~\cite{unet_brain,segresnet} but with $90\%$ fewer parameters. In addition, \MetaSeg significantly outperforms NISF~\cite{nisf}, a related INR approach for segmentation. Ablation results demonstrate that our meta-learning strategy to initialize parameters with joint reconstruction and segmentation supervision is crucial for success.

\MetaSeg offer a new perspective on the capabilities of INRs for imaging tasks, particularly beyond signal representation, which has been their predominant application in the literature. In particular, the results demonstrate that the meta-learning strategy can be useful for image translation tasks such as segmentation, where there is strong cross-correlation across signals. Indeed, as demonstrated by PCA decomposition in Fig.~\ref{fig:pca}, \MetaSeg learns low-dimensional embeddings that seem to jointly model both pixel intensity and semantic structure.

Because an INR explicitly conditions its features on input coordinates, it is likely that an INR will learn highly signal-specific features localized in space. While this is a good property for signal representation, it may lead to poor performance on tasks requiring generalization, such as segmentation. We conducted a pilot test to measure the sensitivity of \MetaSeg to the spatial alignment of test scans to the training set distribution by applying random orientation and translation augmentations. We found that Dice score drops by $2\%$-$6\%$ for test images rotated randomly in $[5\degree,15\degree]$, and drops by $3\%$-$9\%$ for test images with a random translation in $[5-10]$ pixels. Hence, \MetaSeg can be somewhat sensitive to spatial alignment, though further experimentation is needed to understand whether this can be mitigated itself with proper training augmentation.

\MetaSeg learns fast, is resource-friendly, and scales well with more dimensions, unlike typical vision and transformer models. It also is easy to train, using only standard reconstruction and classification losses. In an age where AI requires increasing resource and financial investments, \MetaSeg provides a fresh new perspective to solving image segmentation tasks.

\begin{credits}
\subsubsection{\ackname} The authors acknowledge support from NIH award R01DE032051, ARPA-H award D24AC00296, NSF awards IIS-2107313 and EEC-1648449.
\\
This preprint is pre-peer review and does not include any post-submission improvements or corrections. The final version of record of this contribution is available from the publisher's website with the following \href{https://link.springer.com/chapter/10.1007/978-3-032-04947-6_19}{\footnote{DOI, public URL:\href{https://link.springer.com/chapter/10.1007/978-3-032-04947-6_19}{https://link.springer.com/chapter/10.1007/978-3-032-04947-6\_19}}{DOI and public URL}}.
\end{credits}

%
%
\bibliographystyle{splncs04}
\bibliography{references}

\begin{thebibliography}{10}
\providecommand{\url}[1]{\texttt{#1}}
\providecommand{\urlprefix}{URL }
\providecommand{\doi}[1]{https://doi.org/#1}

\bibitem{inrfusion}
Arefeen, Y., Levac, B., Stoebner, Z., Tamir, J.: Infusion: Diffusion regularized implicit neural representations for 2d and 3d accelerated mri reconstruction. arXiv preprint arXiv:2406.13895  (2024)

\bibitem{spatial_functa}
Bauer, M., Dupont, E., Brock, A., Rosenbaum, D., Schwarz, J.R., Kim, H.: Spatial functa: Scaling functa to imagenet classification and generation (2023), \url{https://arxiv.org/abs/2302.03130}

\bibitem{dice}
Bertels, J., Eelbode, T., Berman, M., Vandermeulen, D., Maes, F., Bisschops, R., Blaschko, M.B.: Optimizing the dice score and jaccard index for medical image segmentation: Theory and practice. In: Medical Image Computing and Computer Assisted Intervention -- MICCAI 2019. pp. 92--100. Springer International Publishing, Cham (2019)

\bibitem{unet_brain}
Buda, M., Saha, A., Mazurowski, M.A.: Association of genomic subtypes of lower-grade gliomas with shape features automatically extracted by a deep learning algorithm. Computers in Biology and Medicine  \textbf{109} (2019). \doi{10.1016/j.compbiomed.2019.05.002}

\bibitem{monai}
Cardoso, M.J., Li, W., Brown, R., Ma, N., Kerfoot, E., Wang, Y., Murray, B., Myronenko, A., Zhao, C., Yang, D., Nath, V., He, Y., Xu, Z., Hatamizadeh, A., Zhu, W., Liu, Y., Zheng, M., Tang, Y., Yang, I., Zephyr, M., Hashemian, B., Alle, S., Zalbagi~Darestani, M., Budd, C., Modat, M., Vercauteren, T., Wang, G., Li, Y., Hu, Y., Fu, Y., Gorman, B., Johnson, H., Genereaux, B., Erdal, B.S., Gupta, V., Diaz-Pinto, A., Dourson, A., Maier-Hein, L., Jaeger, P.F., Baumgartner, M., Kalpathy-Cramer, J., Flores, M., Kirby, J., Cooper, L.A., Roth, H.R., Xu, D., Bericat, D., Floca, R., Zhou, S.K., Shuaib, H., Farahani, K., Maier-Hein, K.H., Aylward, S., Dogra, P., Ourselin, S., Feng, A.: {MONAI: An open-source framework for deep learning in healthcare}  (Nov 2022). \doi{https://doi.org/10.48550/arXiv.2211.02701}

\bibitem{functa}
Dupont, E., Kim, H., Eslami, S.M.A., Rezende, D.J., Rosenbaum, D.: From data to functa: Your data point is a function and you can treat it like one. In: 39th International Conference on Machine Learning (ICML) (2022)

\bibitem{maml}
Finn, C., Abbeel, P., Levine, S.: Model-agnostic meta-learning for fast adaptation of deep networks. In: Precup, D., Teh, Y.W. (eds.) Proceedings of the 34th International Conference on Machine Learning. Proceedings of Machine Learning Research, vol.~70, pp. 1126--1135. PMLR (06--11 Aug 2017), \url{https://proceedings.mlr.press/v70/finn17a.html}

\bibitem{oasis1}
Hoopes, A., Hoffmann, M., Greve, D.N., Fischl, B., Guttag, J., Dalca, A.V.: Learning the effect of registration hyperparameters with hypermorph. Machine Learning for Biomedical Imaging  \textbf{1}(IPMI 2021),  1–30 (Apr 2022). \doi{10.59275/j.melba.2022-74f1}, \url{http://dx.doi.org/10.59275/j.melba.2022-74f1}

\bibitem{vit_medical}
Khan, A., Rauf, Z., Khan, A.R., Rathore, S., Khan, S.H., Shah, N.S., Farooq, U., Asif, H., Asif, A., Zahoora, U., Khalil, R.U., Qamar, S., Asif, U.H., Khan, F.B., Majid, A., Gwak, J.: A recent survey of vision transformers for medical image segmentation (2023), \url{https://arxiv.org/abs/2312.00634}

\bibitem{adam}
Kingma, D.P.: Adam: A method for stochastic optimization. arXiv preprint arXiv:1412.6980  (2014)

\bibitem{focal_loss}
Lin, T.Y., Goyal, P., Girshick, R., He, K., Dollár, P.: Focal loss for dense object detection. IEEE Transactions on Pattern Analysis and Machine Intelligence  \textbf{42}(2),  318--327 (2020). \doi{10.1109/TPAMI.2018.2858826}

\bibitem{oasis2}
Marcus, D.S., Wang, T.H., Parker, J., Csernansky, J.G., Morris, J.C., Buckner, R.L.: {Open Access Series of Imaging Studies (OASIS): Cross-sectional MRI Data in Young, Middle Aged, Nondemented, and Demented Older Adults}. Journal of Cognitive Neuroscience  \textbf{19}(9),  1498--1507 (09 2007). \doi{10.1162/jocn.2007.19.9.1498}, \url{https://doi.org/10.1162/jocn.2007.19.9.1498}

\bibitem{acorn}
Martel, J.N.P., Lindell, D.B., Lin, C.Z., Chan, E.R., Monteiro, M., Wetzstein, G.: Acorn: {Adaptive} coordinate networks for neural scene representation. ACM Trans. Graph. (SIGGRAPH)  \textbf{40}(4) (2021)

\bibitem{segresnet}
Myronenko, A.: 3d mri brain tumor segmentation using autoencoder regularization. In: Crimi, A., Bakas, S., Kuijf, H., Keyvan, F., Reyes, M., van Walsum, T. (eds.) Brainlesion: Glioma, Multiple Sclerosis, Stroke and Traumatic Brain Injuries. pp. 311--320. Springer International Publishing, Cham (2019)

\bibitem{unet}
Ronneberger, O., Fischer, P., Brox, T.: U-net: Convolutional networks for biomedical image segmentation. In: Navab, N., Hornegger, J., Wells, W.M., Frangi, A.F. (eds.) Medical Image Computing and Computer-Assisted Intervention -- MICCAI 2015. pp. 234--241. Springer International Publishing, Cham (2015)

\bibitem{wire}
Saragadam, V., LeJeune, D., Tan, J., Balakrishnan, G., Veeraraghavan, A., Baraniuk, R.G.: Wire: Wavelet implicit neural representations. In: Conf. Computer Vision and Pattern Recognition (2023)

\bibitem{miner}
Saragadam, V., Tan, J., Balakrishnan, G., Baraniuk, R., Veeraraghavan, A.: Miner: Multiscale implicit neural representations. In: European Conf. Computer Vision (2022)

\bibitem{mlr_inr_ct}
Shi, J., Zhu, J., Pelt, D.M., Batenburg, K.J., Blaschko, M.B.: Implicit neural representations for robust joint sparse-view ct reconstruction. arXiv preprint arXiv:2405.02509  (2024)

\bibitem{siren}
Sitzmann, V., Martel, J., Bergman, A., Lindell, D., Wetzstein, G.: Implicit neural representations with periodic activation functions. Advances in neural information processing systems  \textbf{33},  7462--7473 (2020)

\bibitem{nisf}
Stolt-Ans{\'o}, N., McGinnis, J., Pan, J., Hammernik, K., Rueckert, D.: Nisf: Neural implicit segmentation functions. In: Greenspan, H., Madabhushi, A., Mousavi, P., Salcudean, S., Duncan, J., Syeda-Mahmood, T., Taylor, R. (eds.) Medical Image Computing and Computer Assisted Intervention -- MICCAI 2023. pp. 734--744. Springer Nature Switzerland, Cham (2023)

\bibitem{learnit}
Tancik, M., Mildenhall, B., Wang, T., Schmidt, D., Srinivasan, P.P., Barron, J.T., Ng, R.: Learned initializations for optimizing coordinate-based neural representations. In: CVPR (2021)

\bibitem{strainer}
Vyas, K.K., Humayun, I., Dashpute, A., Baraniuk, R., Veeraraghavan, A., Balakrishnan, G.: Learning transferable features for implicit neural representations. In: Advances in Neural Information Processing Systems. vol.~37 (2024)

\bibitem{naf}
Zha, R., Zhang, Y., Li, H.: Naf: Neural attenuation fields for sparse-view cbct reconstruction. In: International Conference on Medical Image Computing and Computer-Assisted Intervention. pp. 442--452. Springer (2022)

\end{thebibliography}

\end{document}